\documentclass[conference]{IEEEtran}
\IEEEoverridecommandlockouts
\usepackage{cite}
\usepackage{amsmath,amssymb,amsfonts}
\usepackage{algorithmic}
\usepackage{graphicx}
\usepackage{textcomp}
\usepackage{xcolor}
\usepackage{stfloats}
\usepackage{booktabs}
\def\BibTeX{{\rm B\kern-.05em{\sc i\kern-.025em b}\kern-.08em
    T\kern-.1667em\lower.7ex\hbox{E}\kern-.125emX}}
\begin{document}

\title{The Neural-Prediction based Acceleration Algorithm of Column Generation for Graph-Based Set Covering Problems \\

\thanks{This work was supported in part by the National Science and Technology Innovation 2030 Major Project of the Ministry of Science and Technology of China under Grant 2018AAA0101604; and in part by the National Natural Science Foundation of China under Grant 61936009. (Corresponding author: Shiji Song.)\par
H. Yuan, P. Jiang and S. Song are with the Department of Automation and BNRist, Tsinghua University, Beijing 100084, China (email: yhf22@mails.tsinghua.edu.cn; jiang-p17@mails.tsinghua.edu.cn; shijis@mail.tsinghua.edu.cn).}
}

\author{
    \IEEEauthorblockN{Haofeng Yuan, Peng Jiang and Shiji Song}
}

\maketitle

\begin{abstract}
Set covering problem is an important class of combinatorial optimization problems, which has been widely applied and studied in many fields. In this paper, we propose an improved column generation algorithm with neural prediction (\textit{CG-P}) for solving graph-based set covering problems. We leverage a graph neural network based neural prediction model to predict the probability to be included in the final solution for each edge. Our \textit{CG-P} algorithm constructs a reduced graph that only contains the edges with higher predicted probability, and this graph reduction process significantly speeds up the solution process. We evaluate the \textit{CG-P} algorithm on railway crew scheduling problems and it outperforms the baseline column generation algorithm. We provide two solution modes for our \textit{CG-P} algorithm. In the optimal mode, we can obtain a solution with an optimality guarantee while reducing the time cost to 63.12\%. In the fast mode, we can obtain a sub-optimal solution with a 7.62\% optimality gap in only 2.91\% computation time.
\end{abstract}

\begin{IEEEkeywords}
set covering problem, column generation, graph neural network, railway crew scheduling problem
\end{IEEEkeywords}

\section{Introduction}
Decades ago, schedules for vehicles, crews, and machines, which can be treated as set covering problems (SCPs), would have been fixed long before the tasks started. But today, due to the real-time information, it is often not the case. The task schedule can change at any time before the start or even during the task. Therefore, the cost of waiting for an optimal scheduling solution becomes more and more expensive. As described above, there is an urgent need for a high-quality solution with minimal computation time for SCPs.
\par
Column generation (CG) algorithm is a powerful approach to solve SCPs, especially large-scale SCPs. CG can keep the problem size manageable. In broad terms, the CG procedure starts with a small subset of columns and iteratively adds new columns if they can potentially improve the current best solution \cite{b1}. The process of adding new columns is often treated as a sub-problem. In this paper, we focus on graph-based set covering problems. Graph-based SCPs are widely applied and studied in many fields, such as vehicle routing, crew scheduling, etc. When solving a graph-based SCP, the sub-problem is always structured as a shortest path problem with resource constraints (SPPRC) \cite{b2}. SPPRC is commonly NP-hard and very time-consuming as the problem size grows. According to \cite{b3}, the solution of the sub-problem consumes more than 90\% of the CPU time in the CG process.
\par
Several accelerating strategies have been proposed for CG, including arc elimination \cite{b7}, problem partitioning \cite{b18}, node treatment \cite{b19}, etc. Based on deep learning techniques, we propose an improved column generation algorithm with neural prediction (\textit{CG-P}) for solving graph-based set covering problems. Given a set of problem instances as training data, we train a neural prediction model based on graph neural network (GNN) to learn the distribution of the optimal solutions on the underlying graph. When solving a new problem instance, we construct a reduced underlying graph that only contains the edges with a higher probability to be included in the final solution. The probability is predicted by the neural prediction model. We evaluate our \textit{CG-P} algorithm on railway crew scheduling problems (RCSPs) and demonstrate that it outperforms the baseline CG algorithm. We provide two solution modes for our \textit{CG-P} algorithm. In the optimal mode, it outperforms the baseline CG in both solution quality and computation time. In the fast mode, a sub-optimal but still high-quality solution can be obtained in an extremely short time. When generalizing to a larger problem with twice the number of edges, our \textit{CG-P} algorithm still performs well. It’s worth mentioning that the neural prediction framework can be easily adapted to other implementations of CG.
\par
The remainder of this paper is organized as follows.  We review the necessary preliminaries in Section \uppercase\expandafter{\romannumeral2}. We introduce the architecture of our neural prediction model in Section \uppercase\expandafter{\romannumeral3}. The \textit{CG-P} algorithm is introduced in Section \uppercase\expandafter{\romannumeral4}. Experiment results on RCSP are presented in Section \uppercase\expandafter{\romannumeral5} and Section 
\uppercase\expandafter{\romannumeral6} briefly concludes.

\section{Preliminaries}

\subsection{Set Covering Problems}
Given a set of elements $\left \{ 1,2,\cdots ,m \right \}$, called \textit{universe} set, and a collection $N$ of $n$ \textit{subsets}, a set covering problem is to identify a sub-collection $S \subseteq N$  at minimal cost such that the union of the sub-collection should equal the universe.
\par
SCP can formally be defined as follows. Let $A=\left[a_{ij}\right]_{m \times n} $ be a 0-1 $m\times n$ matrix, and $c=\left ( c_{1},\cdots,c_{n} \right)$ be a $n$-dimensional cost vector. We will refer to the rows and columns of $A$ simply as rows $M=\left \{ 1,\cdots,m \right \} $ and columns $N=\left \{ 1,\cdots,n \right \} $. SCP is the problem of covering the rows of $A$ by a subset of columns at a minimal cost. The value $c_{j} \left( j \in N \right)$ represents the cost of column $j$, assuming $c_{j}>0$ for $ j \in N$ without loss of generality. We say that $ a_{ij}=1$ if row $ i \in M$ is covered by column $ j \in N$ or $a_{ij}=0$ otherwise. A natural mathematical model for SCP is: 
\begin{align}
\min_{x} &\sum_{j \in N} c_{j} x_{j} \tag{1}\\
{\rm s.}{\rm t.}\,&\sum_{j \in N} a_{i j} x_{j} \geq 1,  i \in M \tag{2}\\
&x_{j} \in\{0,1\} , j \in N \tag{3}
\end{align}where $x_{j}=1$ if $j \in S$, or $x_{j}=0$ otherwise. Eq. $\left( 2 \right)$ ensures that each row is covered by at least one column and $\left( 3 \right)$ represents the integrality constraints  \cite{b4}.
\par
In this paper, we focus on graph-based SCPs. The universe set and subsets are defined on an underlying graph. For example, the universe set can be the nodes on the full graph, and a subset can be the nodes along a path.

\subsection{Column Generation Algorithm}
Column generation is an efficient approach to solving linear programs (LP) with a large number of columns. A large number of columns means a large number of potential subsets, and it is common in SCPs. It is impractical to consider all the columns explicitly in an LP. To address this issue, the CG procedure starts with a restricted problem with only a small part of columns. Other feasible columns are iteratively added to the restricted problem if they are potential to improve the objective value. Once we can demonstrate that any new columns would no longer improve the objective value, the CG procedure stops \cite{b1}.\par
To solve the whole problem, called \textit{master problem} (MP), the CG algorithm iteratively solves a \textit{restricted master problem} (RMP) and a \textit{sub-problem} (SP). The sub-problem is also called the pricing problem. In RMP, we need to solve a LP problem with a subset of columns, and provide dual variable $u=\left ( u_{1},\cdots,u_{m} \right)$ for the sub-problem. Each dimension of the dual variable corresponds to a row of $A$. The \textit{reduced cost} $\bar{c}_{j}$ of column $j$ is given by:
\begin{equation}
\bar{c}_{j}:=c_{j}-\sum_{i \in M}{u}_{i}{a}_{ij} \tag{4}
\end{equation}\par
The role of SP is a column generator to provide columns that are potential to improve the objective value, or to prove that none exists. In SP, we need to find potential columns with minimal negative reduced cost. Otherwise, if all potential columns have a non-negative reduced cost, we can demonstrate that adding any new column would no longer improve the objective value, then the CG procedure stops. To achieve an integer solution, some extra steps are required to handle the integrality constraints.

\subsection{Railway Crew Scheduling Problems}
A railway crew scheduling problem consists of finding a set of anonymous \textit{duties} at minimum cost that covers all \textit{trips} in a period of time, e.g., a single workday \cite{b5}. A \textit{trip} is a task unit that has to be carried out by a crew member. A sequence of trips performed by a crew member forms a \textit{duty}. The feasibility of duty is restricted by many rules and regulations, such as geographical and chronological constraints between consecutive trips, a maximum total working time, etc.
\par
RCSP can be formed as a graph-based set covering problem. The universe set is composed of all the trips to be covered, and each subset is a feasible duty. A RCSP instance corresponds to a time-space network as the underlying graph. Each node represents a trip and each edge represents a feasible connection. Fig. 1 is an example of the time-space network of RCSP. A feasible path satisfying the connection constraints and resource constraints forms a duty. Since RCSPs are usually large-scale, the vast majority of researchers use column generation techniques to solve them \cite{b5}.
\begin{figure}[t]
\centering
\includegraphics[width=3.5in, keepaspectratio]{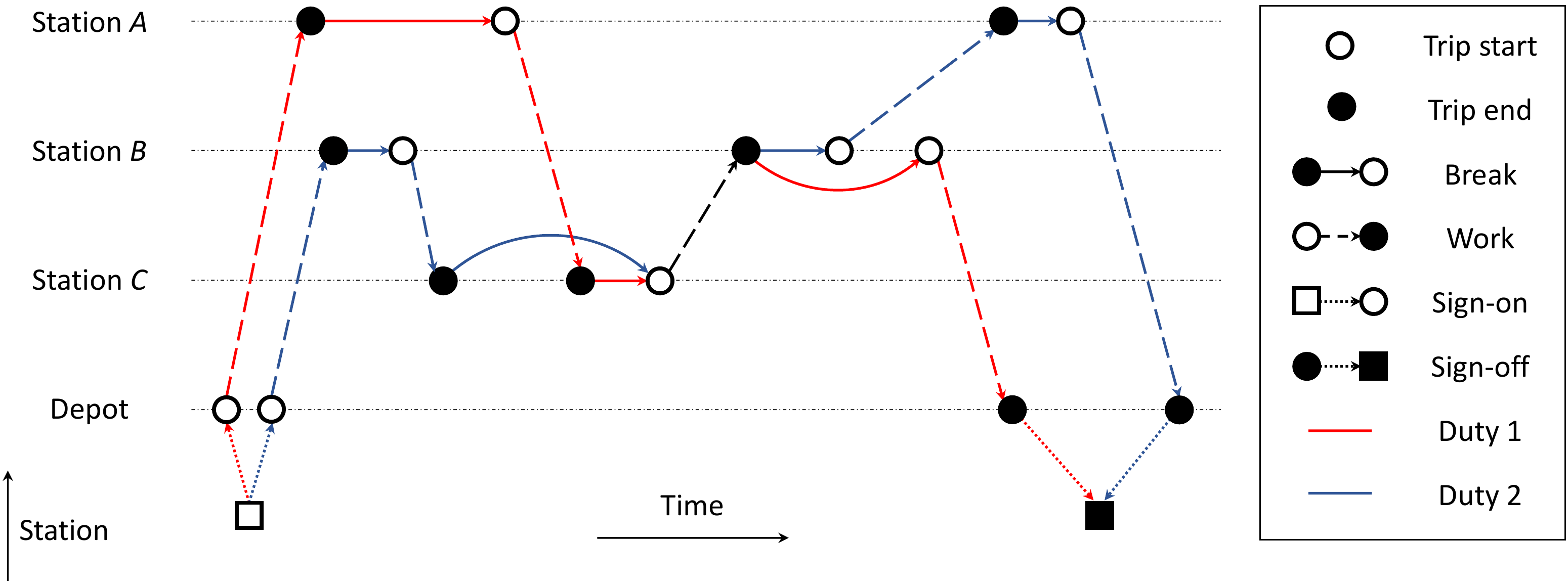}\\
\caption{A time-space network of RCSP.}
\end{figure}
\section{Neural Prediction Model}
\par
When solving a graph-based SCP, the sub-problem is always constructed as a SPPRC. Each column in the master problem represents a feasible path on the underlying graph. For a problem instance and its underlying graph $G=(V,E)$, we can define an edge subset $E^{\prime} \subseteq E$ that only contains the edges included in the columns with non-zero coefficient in the final solution. That is, the subset $E^{\prime}$ only contains the edges selected in the final solution. Those edges are called \textit{valid edges}. 
\par
Our idea is to predict the probability that an edge in $E$ belongs to the subset $E^{\prime}$ based on a deep learning model. When solving a new problem instance, we can predict this probability for edges on the underlying graph, and construct a reduced graph that only contains the edges with a higher probability to be included in the valid edge set. Essentially, it is an edge classification task on the graph.
\par
Graph neural networks are neural models operating on graph domain to capture the dependence of graphs via message passing between nodes and edges \cite{b6}. GNN-based techniques have been proposed for many graph-based combinatorial optimization problems such as traveling salesman problem \cite{b8}, maximum cut problem \cite{b9}, minimum vertex cover problem \cite{b10}, etc. Recent advances in graph neural network techniques are a good fit for our task, because GNN naturally operate on the graph structure.
\par
Given a graph as input, we train a GNN-based model to output a score for every edge, representing the probability of being included in the final solution. The model architecture is shown in Fig. 2. The input graph is passed through the neural network and linked to the ground-truth valid edge set so that the model parameters can be trained end-to-end by minimizing the cross-entropy loss via gradient descent.

\begin{figure*}[ht]
\centering
\includegraphics[width=7in, keepaspectratio]{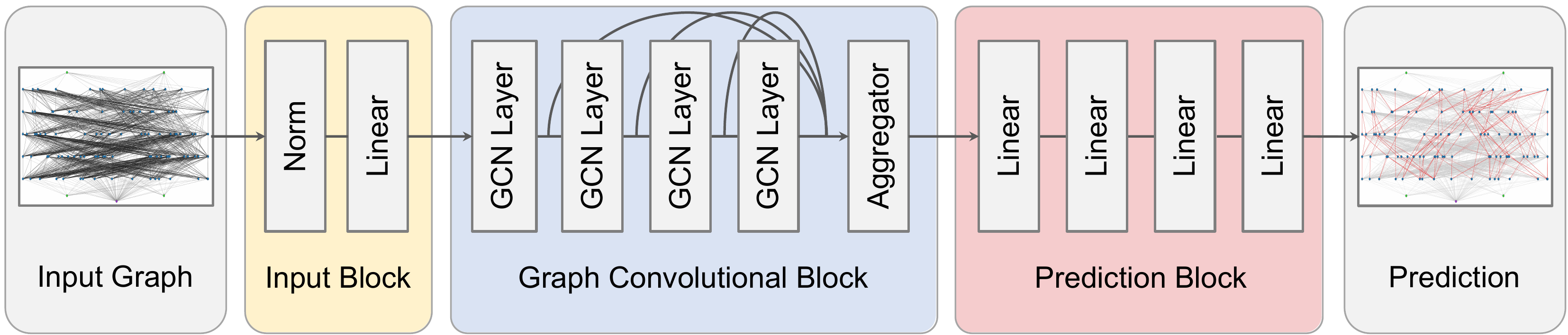}\\
\caption{Architecture of the neural prediction model}
\end{figure*}

\subsection{Input Block}
For different problems, the scales between dimensions of the input features can be greatly different. It is unfavorable for the training of deep learning models. Therefore, every dimension of the input node features $x_{i}^{init}$ and edge features $e_{ij}^{init}$ is normalized to $\left [0,1 \right ] $. Then the normalized features $x_{i}$ and $e_{ij}$ are embedded to high-dimensional embeddings through a linear projection followed by batch normalization (BN) \cite{b20}:
\begin{gather}
x_{i}=\operatorname{Norm}\left(x_{i}^{init}\right) \tag{5} \\
e_{ij}=\operatorname{Norm}\left(e_{ij}^{init}\right) \tag{6} \\
x_{i}^{0}=\operatorname{BN}\left(W_{1} x_{i}+b_{1}\right) \tag{7} \\
e_{ij}^{0}=\operatorname{BN}\left(W_{2} e_{ij}+b_{2}\right) \tag{8}
\end{gather}
\subsection{Graph Convolutional Block}

The embeddings are updated through $l_{conv}$  message passing layers. Inspired by the experiments in \cite{b11} and \cite{b12}, we apply residual GatedGCN layers with additional edge feature representations and a dense attention map $\eta_{ij}$, which makes the diffusion process anisotropic on the graph. The forward pass process is given by:
\begin{gather}
{\hat{e}}_{ij}^\ell=A^\ell x_i^\ell+B^\ell x_j^\ell+C^\ell e_{ij}^\ell \tag{9} \\
\eta_{ij}^{\ell}=\frac{\sigma\left(\hat{e}_{ij}^{\ell+1}\right)}{\sum_{j^{\prime} \sim i}\sigma\left(\hat{e}_{ij^{\prime}}^{\ell+1}\right)} \tag{10}\\
\hat{x}_{i}^{\ell}=U^{\ell} x_{i}^{\ell}+\sum_{j \sim i} \eta_{i j}^{\ell} \odot V^{\ell} x_{j}^{\ell} \tag{11}\\
e_{ij}^{\ell+1}=e_{ij}^{\ell}+\operatorname{ReLU}\left(\operatorname{BN}\left(\hat{e}_{ij}^{\ell}\right)\right) \tag{12}\\
x_{i}^{\ell+1}=x_{i}^{\ell}+\operatorname{ReLU}\left(\operatorname{BN}\left(\hat{x}_{i}^{\ell}\right)\right) \tag{13}
\end{gather}where $x_i^\ell$ and $e_{ij}^\ell$ are the output node and edge embeddings of layer $\ell$, and $A^\ell$, $B^\ell$, $C^\ell$, $U^\ell$, $V^\ell$ are learnable matrices.
\par
Additionally, to improve the insensitivity to different graph sizes and block depth, we apply a jump connection structure to edge embeddings \cite{b13}, with an element-wise max-pooling aggregation to select the most informative layer for each embedding coordinate. The aggregated edge embeddings $e_{ij}^{final}$ are given as follows:
\begin{equation}
e_{ij}^{final}=\operatorname{Maxpool}\left(e_{i j}^{0}, e_{i j}^{1}, \cdots, e_{i j}^{l_{conv}}\right) \tag{14}
\end{equation}
\subsection{Prediction Block}
Finally, the final edge embeddings $e_{ij}^{final}$ are used to compute the probability score $p_{ij}$ that edge $ij$ is included in the valid edge set. This score can be visualized as a probability heat-map over the graph, as shown in Fig. 4(c). Each score $p_{ij}\in\left[0,1\right]$ is given by a multi-layer perceptron (MLP) with $l_{mlp}$ layers and end with a sigmoid output layer:
\begin{equation}
p_{ij}=\operatorname{MLP}\left(e_{ij}^{final} \right) \tag{14}
\end{equation}\par
Given the ground-truth valid edge set, we minimize a weighted binary cross-entropy loss over mini-batches. As the graph size increases, the classification task becomes highly unbalanced towards the negative class. Therefore, an appropriate class weight is required to balance this effect.

\section{Column Generation with Neural Prediction}
\begin{figure}[b]
\centering
\includegraphics[width=2.8in, keepaspectratio]{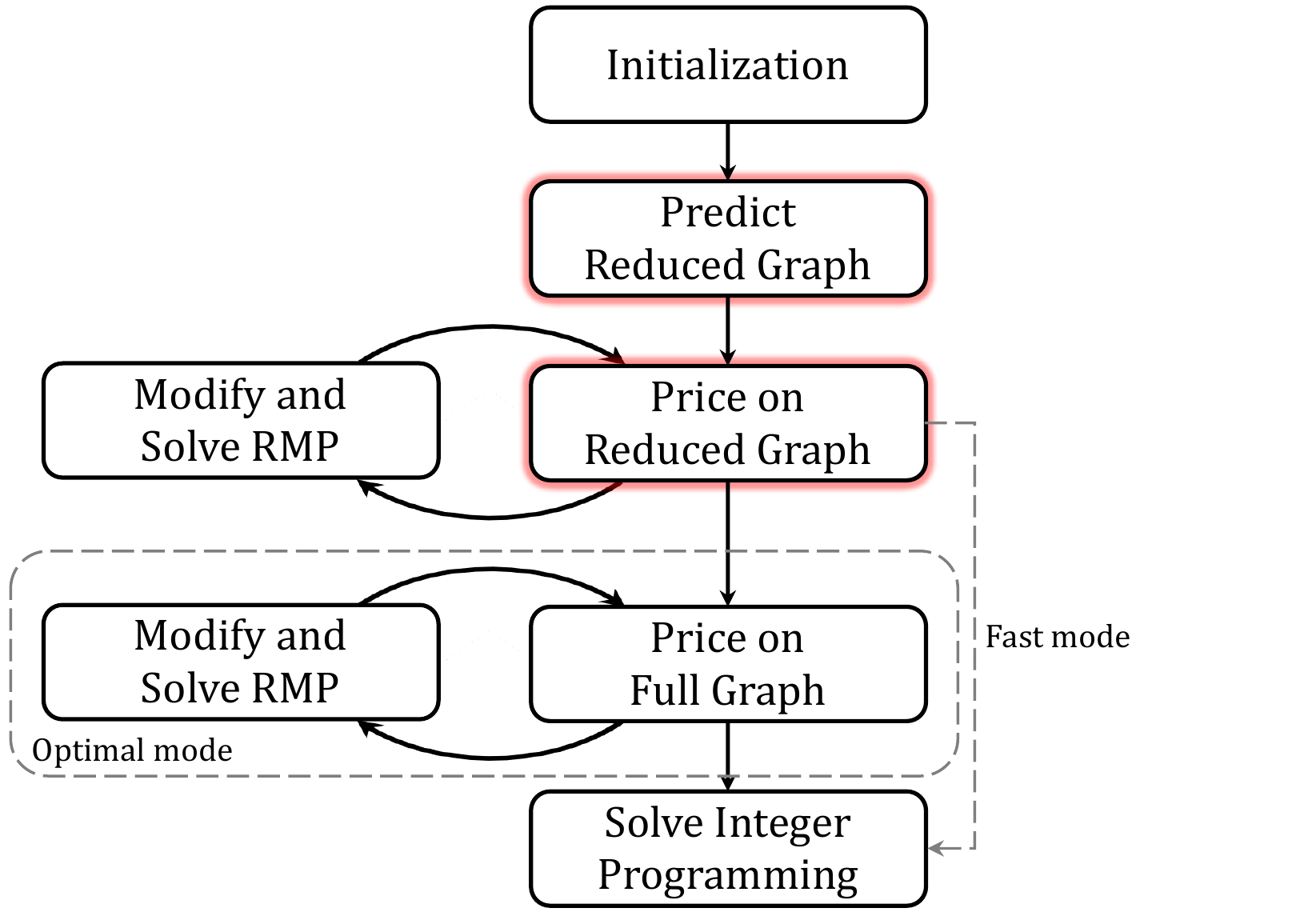}\\
\caption{Process of the \textit{CG-P} algorithm. The optimal mode means pricing on the reduced graph and full graph successively, and the fast mode means pricing only on the reduced graph.}
\end{figure}
Our \textit{CG-P} algorithm is based on the CG framework. The key improvement of our \textit{CG-P} is to add a prediction step to reduce the size of sub-problem. Instead of always pricing on the full underlying graph like the original CG, we use our neural prediction model to predict for the edges on the underlying graph, and construct a reduced graph that only contains the edges with higher probability to be included in the final solution, as shown in Fig. 4(d). Pricing on this much simpler but high-quality graph can significantly speed up the solution process. The process of \textit{CG-P} algorithm is shown in Fig. 3. We will introduce the details in this section.
\begin{figure*}[ht]
\centering
\includegraphics[width=7in, keepaspectratio]{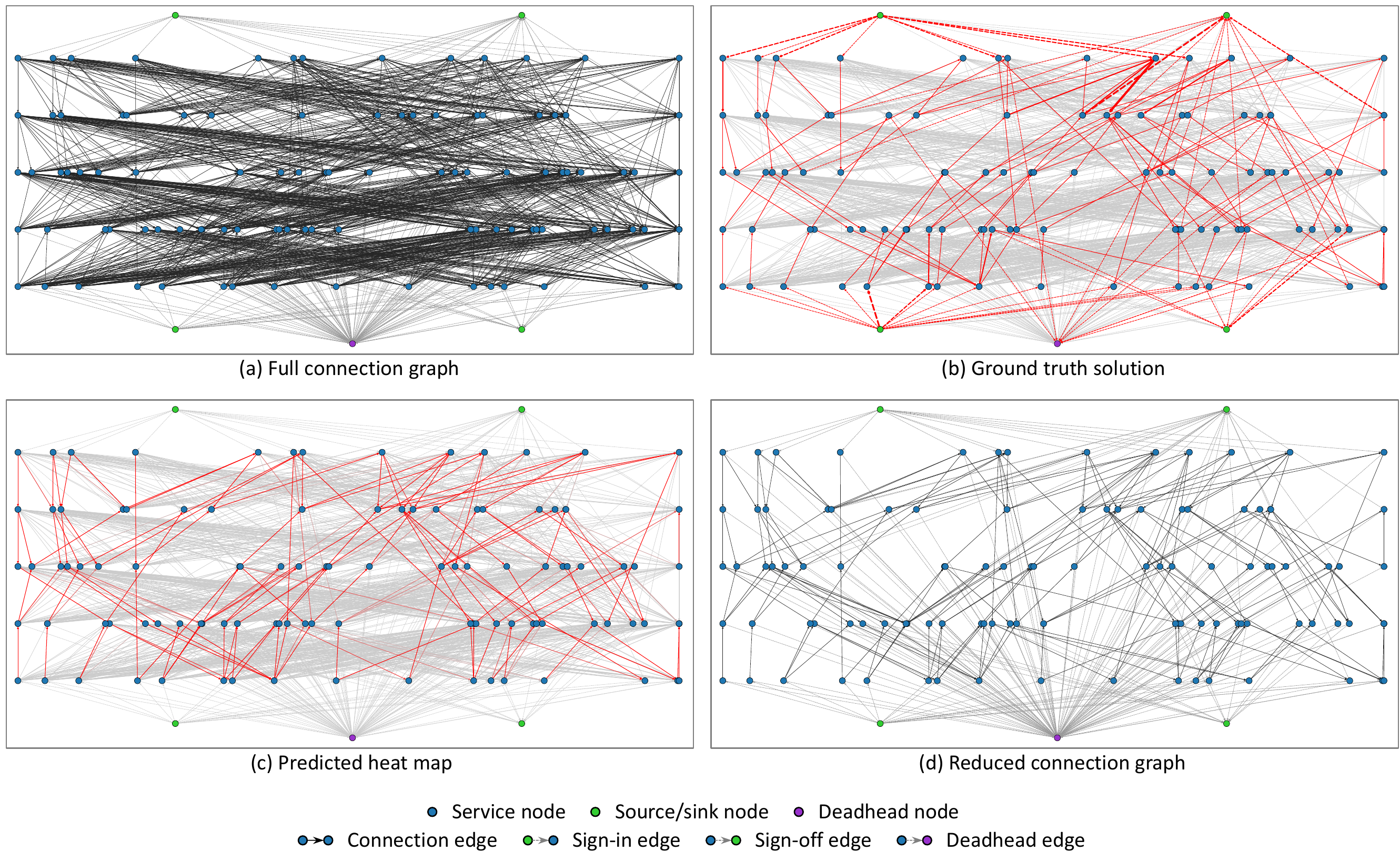}\\
\caption{Neural prediction and graph reduction in our experiment.}
\end{figure*}
\subsection{Column Generation Framework}
Linear relaxation is usually applied to the RMP. In most researches, the solution is obtained by LP methods of a commercial solver \cite{b14}. We choose \textit{Gurobi} as our LP solver.
\par
The sub-problem is solved based on the dual variable obtained from RMP. For graph-based set covering problems, the SP is commonly structured as a SPPRC and solved by dynamic programming (DP) based approaches. Based on the DP framework,  we employed a label setting algorithm proposed in \cite{b15}. We would like to mention that our neural prediction framework is applicable regardless of the specific dynamic programming approach used.
\par
Since the solution of relaxed MP does not necessarily satisfy the integrality constraints, some extra steps are required to generate the integer solution. We solve an integer programming (IP) in a subsequent step to obtain the integer solution using a commercial solver after the relaxed MP is solved to optimality. The final RMP is treated as an IP and solved with \textit{Gurobi}. Some researchers applied other IP methods like branch-and-price \cite{b16}, but as mentioned above, the specific methods applied make no difference to our prediction framework.
\subsection{Pricing on Reduced Graph}
Based on the CG framework, we add a prediction and reduction step. We construct a reduced graph that only contains those edges with a higher probability to be included in the final solution using the neural prediction model. As in Fig. 3, in the early stage, we only price on the reduced graph. It allows us to generate high-quality columns even lacking sufficient price information. Once the RMP solution cannot be improved on the reduced graph, we switch to price on the full underlying graph until the relaxed MP is solved to optimality. This process does not affect the optimality of the relaxed MP solution, and our experiment shows that the quality of integer solutions can even be slightly improved. We call this process of pricing on the reduced graph and the full graph successively to obtain the optimal solution of relaxed MP \textit{optimal mode}. For further reduction of the computation time, we can price only on the reduced graph and stop the CG process immediately if the RMP solution cannot be improved on the reduced graph, called \textit{fast mode}. Thus, a sub-optimal but still high-quality solution can be obtained in an extremely short time.

\section{Experiment}
\begin{figure*}[ht]
\centering
\includegraphics[width=7in, keepaspectratio]{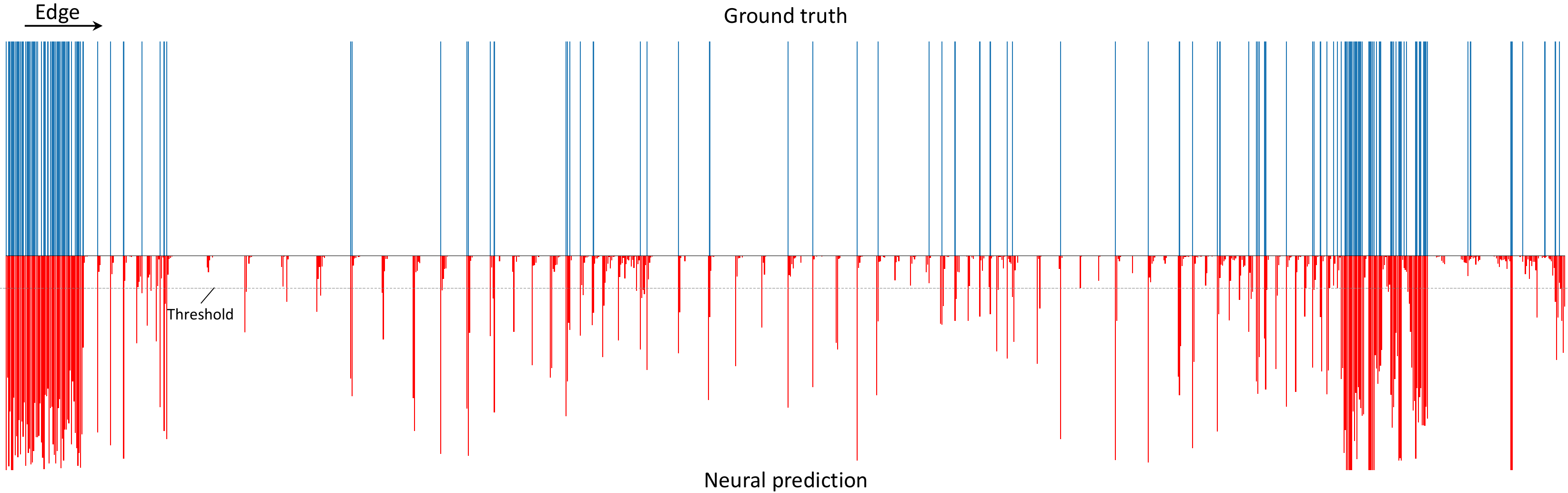}\\
\caption{Neural prediction results of an experiment instance. The horizontal axis represents the edges. The upper part of the figure represents the ground-truth valid edge set, while the lower part represents the probability predicted by our neural prediction model. We select the edges with a predicted probability greater than the threshold for the reduced graph.}
\end{figure*}
We experiment on railway crew scheduling problems, a typical class of graph-based set covering problem, to evaluate the performance of our \textit{CG-P} algorithm.
\subsection{Setup}
Our problem instances are generated following the train timetabling on a railway line with 5 stations. Train schedules, as well as necessary attributes, are randomly generated, including the start and end stations, intermediate stops, train timetables, etc. Each duty is restricted to a maximum working time of 8 hours. In addition, the first train is fixed for all instances to ensure feasibility. We define the objective function as to minimize the sum of total duties with extra cost for deadhead. Deadhead means away from the crew bases at end of the duty.
\par
The underlying graph contains four types of nodes and four types of edges. For each crew base, we introduce a \textit{source node} and a \textit{sink node}. Each trip corresponds to a \textit{service node}. The \textit{deadhead node} indicates the end of a duty that does not end at a crew base. A duty must start from a source node and end at a sink/deadhead node. \textit{Sign-in edges} are introduced from each source node to service nodes from the crew base and similarly \textit{sign-off edges}. From each service node not arriving at a crew base, a \textit{deadhead edge} is set to the deadhead node. We set \textit{connection edges} for all service node pairs satisfying the geographical constraint. Additionally, a chronological constraint of at least 15 minutes of transit time should be guaranteed for a pair. To handle the working time constraint, we set a time resource on each edge. We set the cost of sign-off edges to 1 and the deadhead edge to 1.5 to indicate the extra cost for deadhead. The cost of other edges depends on the dual variable provided by RMP.

\subsection{Model Configuration}

We generated 9,000 RCSP instances following the rules described above, containing about 10 million edges totally. Each instance contains 80-140 nodes (average 110) and 700-2000 edges (average 1200). We collect the solutions and valid edge sets of these problem instances through the baseline CG algorithm (original CG without prediction) as the training data for our neural prediction model.
\par 
We build the neural prediction model as described in Section \uppercase\expandafter{\romannumeral3}. The graph convolutional block consists of $l_{conv}=4$ layers with hidden dimension $h_{conv}=256$ for each embedding. The prediction block consists of $l_{mlp}=4$ layers in the MLP with hidden dimension $h_{mlp}=256$ for each layer. To fit our problem size, we set $w=0.15$ for the weight of negative samples in our loss function to address the high unbalance towards the negative class.
\par
We follow a standard training procedure to train our neural prediction model. We train the prediction model to directly output a score for all connection edges representing the probability of being included in the valid edge set by minimizing the cross-entropy loss via gradient descent. The training set is divided into mini-batches of 10 graphs each. We use the Adam optimizer and set the learning rate to 3e-4. The training process takes around 150 epochs to achieve the best validation loss on a single NVIDIA Tesla P100 GPU, taking about 2 hours. In the \textit{CG-P} algorithm, we select the edges with a predicted probability higher than 0.15 for the reduced graph. The predicted probability is shown in Fig. 5.

\subsection{Results}

We evaluate the solution quality and computation time on 100 unseen problem instances. We compare our \textit{CG-P} algorithm to the baseline CG and a CG algorithm that solves the sub-problem through a genetic algorithm (\textit{CG-GA}) \cite{b17}.
\par

The performance of different algorithms is shown in Table \uppercase\expandafter{\romannumeral1}. The baseline CG provides an optimal solution to the relaxed MP. However, It consumes a considerable long considerably computation time. \textit{CG-GA} reduces the computation time but cannot guarantee optimality. It is an inherent defect of heuristic algorithms including the genetic algorithm. Furthermore, the trade-off between solution quality and computation time is often troublesome for researchers. In contrast, the optimal mode of our \textit{CG-P} algorithm outperforms the baseline CG and \textit{CG-GA} in both solution quality and computation time. Consuming only 63.12\% of computation time, we can obtain a solution with an optimality guarantee to the relaxed MP. The final IP solution is even better than baseline CG in our evaluation. Furthermore, if we require a solution in an extremely short time and are accept a little loss of optimality, we can choose the fast mode to obtain a sub-optimal solution with a 7.62\% optimality gap in only 2.91\% time. In addition, we also evaluate the generalization performance of \textit{CG-P} on a problem instance with 2400 edges, twice the training set, and obtained a similarly good result.
\par
We can also demonstrate the advantages of our \textit{CG-P} algorithm through the optimization process of RMP objective value, as Fig. 6. Compared to the baseline CG and \textit{CG-GA}, the reduced graph can provide high-quality columns even in the early stage, significantly speeding up the solution process.

\begin{figure}[t]
\centering
\includegraphics[width=3.2in, keepaspectratio]{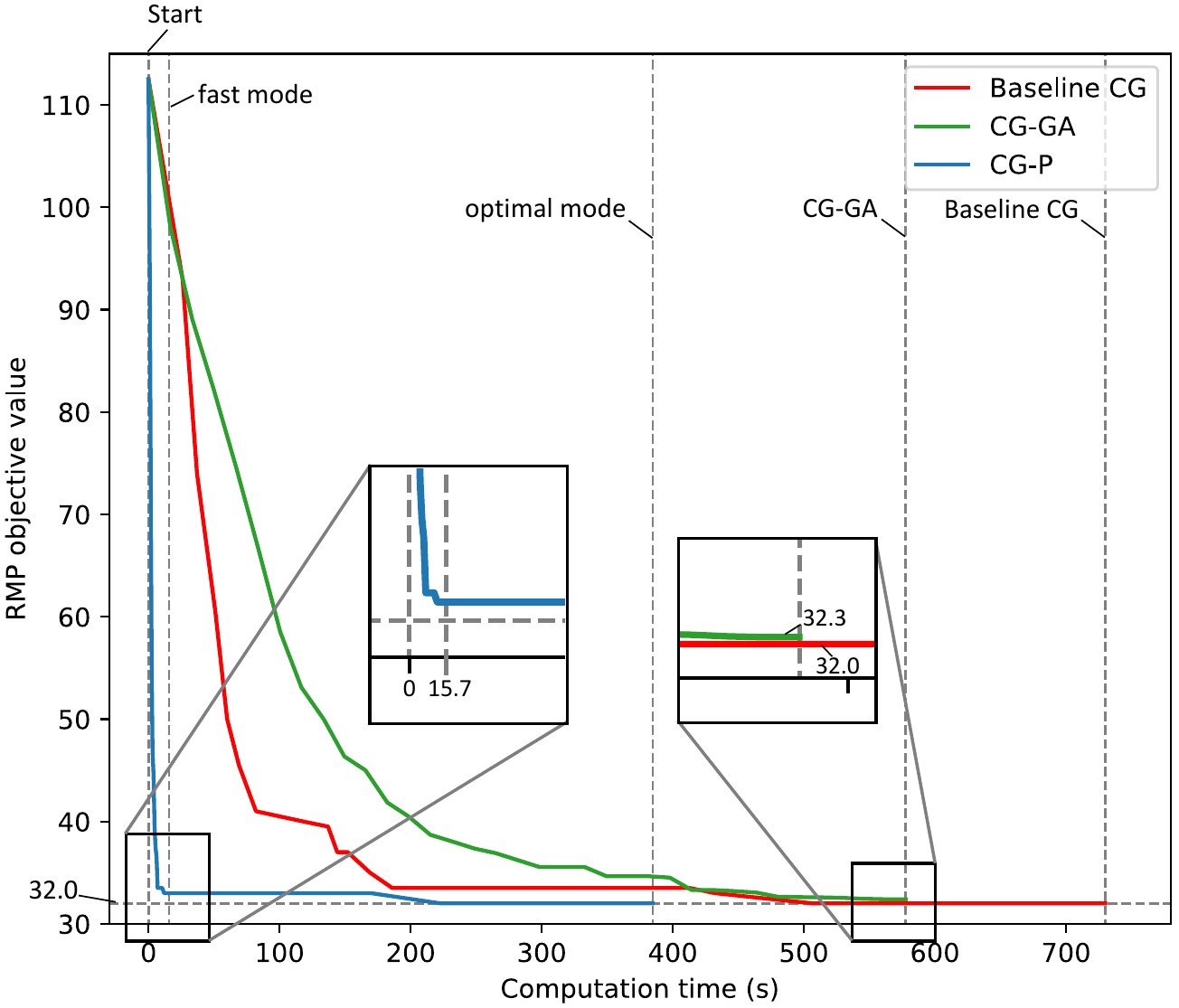}\\
\caption{Comparison of the optimization process of RMP objective value through different CG algorithms. The vertical dashed lines represent the termination time of different algorithms.}
\end{figure}

\begin{table}[t]
\renewcommand\arraystretch{1.1}
\caption{Result Comparison of Different CG Algorithms}
\begin{center}
\begin{tabular}{l|cc|cc}
\toprule
\textbf{Algorithm} & \textbf{Solution} & \textbf{Gap(\%)} & \textbf{Time(s)} & \textbf{Ratio(\%)}        \\
\midrule
Baseline CG & 29.72 & \textbackslash{} & 679.53 & \textbackslash{} \\
\textit{CG-GA} \cite{b17}   & 30.11    & 1.31     & 595.18  & 87.59 \\
\textbf{\emph{CG-P} (optimal)} & \textbf{29.71} & \textbf{-0.04} & 428.92 & 63.12   \\
\textbf{\emph{CG-P} (fast)}        & 31.99    & 7.62     & \textbf{19.76} & \textbf{2.91} \\
\bottomrule
\end{tabular}
\label{tab1}
\end{center}
\end{table}

\section{Conclusions}
In this paper, we propose an improved column generation algorithm with neural prediction (\textit{CG-P}) for solving graph-based set covering problems. We leverage a neural prediction model to predict the probability to be included in the final solution for each edge, and construct a reduced graph that only contains edges with higher predicted probability. Our \textit{CG-P} outperforms the baseline CG and \textit{CG-GA}  in both computation time and solution quality. Future works shall explore incorporating reinforcement learning and transfer learning into our framework to handle larger-scale and wider-range problems.

\bibliographystyle{IEEEtran}
\bibliography{IEEEabrv,reference}

\begin{thebibliography}{10}
\providecommand{\url}[1]{#1}
\csname url@samestyle\endcsname
\providecommand{\newblock}{\relax}
\providecommand{\bibinfo}[2]{#2}
\providecommand{\BIBentrySTDinterwordspacing}{\spaceskip=0pt\relax}
\providecommand{\BIBentryALTinterwordstretchfactor}{4}
\providecommand{\BIBentryALTinterwordspacing}{\spaceskip=\fontdimen2\font plus
\BIBentryALTinterwordstretchfactor\fontdimen3\font minus
  \fontdimen4\font\relax}
\providecommand{\BIBforeignlanguage}[2]{{%
\expandafter\ifx\csname l@#1\endcsname\relax
\typeout{** WARNING: IEEEtran.bst: No hyphenation pattern has been}%
\typeout{** loaded for the language `#1'. Using the pattern for}%
\typeout{** the default language instead.}%
\else
\language=\csname l@#1\endcsname
\fi
#2}}
\providecommand{\BIBdecl}{\relax}
\BIBdecl

\bibitem{b1}
M.~E. L{\"u}bbecke and J.~Desrosiers, ``Selected topics in column generation,''
  \emph{Operations research}, vol.~53, no.~6, pp. 1007--1023, 2005.

\bibitem{b2}
S.~Irnich and G.~Desaulniers, ``Shortest path problems with resource
  constraints,'' in \emph{Column generation}.\hskip 1em plus 0.5em minus
  0.4em\relax Springer, 2005, pp. 33--65.

\bibitem{b3}
R.~V{\'{a}}clav{\'{\i}}k, A.~Nov{\'{a}}k, P.~Sucha, and Z.~Hanz{\'{a}}lek,
  ``Accelerating the branch-and-price algorithm using machine learning,''
  \emph{European Journal of Operational Research}, vol. 271, no.~3, pp.
  1055--1069, 2018.

\bibitem{b7}
G.~Desaulniers, J.~Desrosiers, Y.~Dumas, M.~M. Solomon, and F.~Soumis, ``Daily
  aircraft routing and scheduling,'' \emph{Management Science}, vol.~43, no.~6,
  pp. 841--855, 1997.

\bibitem{b18}
M.~Gamache, F.~Soumis, G.~Marquis, and J.~Desrosiers, ``A column generation
  approach for large-scale aircrew rostering problems,'' \emph{Operations
  research}, vol.~47, no.~2, pp. 247--263, 1999.

\bibitem{b19}
J.~Desrosiers, Y.~Dumas, M.~M. Solomon, and F.~Soumis, ``Time constrained
  routing and scheduling,'' \emph{Handbooks in operations research and
  management science}, vol.~8, pp. 35--139, 1995.

\bibitem{b4}
A.~Caprara, P.~Toth, and M.~Fischetti, ``Algorithms for the set covering
  problem,'' \emph{Annals of Operations Research}, vol.~98, no.~1, pp.
  353--371, 2000.

\bibitem{b5}
J.~Heil, K.~Hoffmann, and U.~Buscher, ``Railway crew scheduling: Models,
  methods and applications,'' \emph{European journal of operational research},
  vol. 283, no.~2, pp. 405--425, 2020.

\bibitem{b6}
J.~Zhou, G.~Cui, S.~Hu, Z.~Zhang, C.~Yang, Z.~Liu, L.~Wang, C.~Li, and M.~Sun,
  ``Graph neural networks: A review of methods and applications,'' \emph{AI
  Open}, vol.~1, pp. 57--81, 2020.

\bibitem{b8}
B.~Hudson, Q.~Li, M.~Malencia, and A.~Prorok, ``Graph neural network guided
  local search for the traveling salesperson problem,'' \emph{arXiv preprint
  arXiv:2110.05291}, 2021.

\bibitem{b9}
S.~Gu and Y.~Yang, ``A deep learning algorithm for the max-cut problem based on
  pointer network structure with supervised learning and reinforcement learning
  strategies,'' \emph{Mathematics}, vol.~8, no.~2, p. 298, 2020.

\bibitem{b10}
E.~Khalil, H.~Dai, Y.~Zhang, B.~Dilkina, and L.~Song, ``Learning combinatorial
  optimization algorithms over graphs,'' \emph{Advances in neural information
  processing systems}, vol.~30, 2017.

\bibitem{b20}
S.~Ioffe and C.~Szegedy, ``Batch normalization: Accelerating deep network
  training by reducing internal covariate shift,'' in \emph{International
  conference on machine learning}.\hskip 1em plus 0.5em minus 0.4em\relax PMLR,
  2015, pp. 448--456.

\bibitem{b11}
C.~K. Joshi, T.~Laurent, and X.~Bresson, ``An efficient graph convolutional
  network technique for the travelling salesman problem,'' \emph{arXiv preprint
  arXiv:1906.01227}, 2019.

\bibitem{b12}
V.~P. Dwivedi, C.~K. Joshi, T.~Laurent, Y.~Bengio, and X.~Bresson,
  ``Benchmarking graph neural networks,'' \emph{arXiv preprint
  arXiv:2003.00982}, 2020.

\bibitem{b13}
K.~Xu, C.~Li, Y.~Tian, T.~Sonobe, K.-i. Kawarabayashi, and S.~Jegelka,
  ``Representation learning on graphs with jumping knowledge networks,'' in
  \emph{International Conference on Machine Learning}.\hskip 1em plus 0.5em
  minus 0.4em\relax PMLR, 2018, pp. 5453--5462.

\bibitem{b14}
H.~D. Mittelmann, ``Latest benchmarks of optimization software,'' in
  \emph{INFORMS Annual Meeting 2017}, 2017.

\bibitem{b15}
M.~Desrochers and F.~Soumis, ``A generalized permanent labelling algorithm for
  the shortest path problem with time windows,'' \emph{INFOR: Information
  Systems and Operational Research}, vol.~26, no.~3, pp. 191--212, 1988.

\bibitem{b16}
C.~Barnhart, E.~L. Johnson, G.~L. Nemhauser, M.~W. Savelsbergh, and P.~H.
  Vance, ``Branch-and-price: Column generation for solving huge integer
  programs,'' \emph{Operations research}, vol.~46, no.~3, pp. 316--329, 1998.

\bibitem{b17}
K.~Hoffmann, U.~Buscher, J.~S. Neufeld, and F.~Tamke, ``Solving practical
  railway crew scheduling problems with attendance rates,'' \emph{Business \&
  Information Systems Engineering}, vol.~59, no.~3, pp. 147--159, 2017.

\end{thebibliography}

\end{document}